\documentclass[runningheads]{llncs}
\usepackage[T1]{fontenc}
\usepackage{graphicx}
\usepackage{booktabs}
\usepackage[misc]{ifsym}
\newcommand{\corr}{(\Letter)}

\usepackage{amsmath}
\usepackage{dsfont}
\usepackage{url}

\begin{document}

\title{Geometric Attractor Monitoring: A Robust and Frugal Framework for Multi-modal Industrial Robotic Cycles}

\titlerunning{Geometric Attractor Monitoring}

\author{Martin Bonsergent-Brachet\inst{1}\corr \and Jesse Read\inst{1} \and Dany Abboud\inst{2}}

\authorrunning{M. Bonsergent-Brachet et al.}

\toctitle{Geometric Attractor Monitoring: A Robust and Frugal Framework for Multi-modal Industrial Robotic Cycles}
\tocauthor{Martin Bonsergent-Brachet, Jesse Read, Dany Abboud}

\institute{LIX, École Polytechnique, Institut Polytechnique de Paris, Palaiseau, France\\
\email{martin.bonsergent-brachet@lix.polytechnique.fr}, \email{jesse.read@polytechnique.edu}
\and Renault, France\\
\email{dany.abboud@renault.com}}

\maketitle

\begin{abstract}
Monitoring the health of heterogeneous industrial robot fleets is severely challenged by the multi-modal nature of their operational cycles and a persistent scarcity of run-to-failure data. Standard data-driven approaches, particularly deep learning architectures relying on sequential reconstruction, often struggle in this specific setting; they tend to over-smooth complex dynamics, masking early signs of degradation. To address these industrial constraints, we reframe the monitoring problem through a framework based on Phase Space Reconstruction (PSR). Instead of predicting temporal sequences, this framework transforms univariate sensor data into a geometric attractor, explicitly unfolding the mechanical states independently of their temporal occurrence. By evaluating various anomaly scoring techniques within this space, we demonstrate that discrete support estimation provides an effective and computationally frugal Health Indicator (HI). Validated on a real-world dataset of 21 heterogeneous robots over three years and a synthetic Langevin system, our approach outperforms standard deep learning baselines. We show that aligning the algorithmic bias with the geometric properties of the target system yields a pragmatic, traceable and easily deployable approach perfectly tailored to the realities of industrial constraints.

\end{abstract}

\section{Introduction}

In the era of Industry 4.0, industrial automation has become the cornerstone of modern manufacturing, particularly in high-throughput sectors like the automotive industry. Production lines rely on vast fleets of industrial robots operating continuously to meet demanding production targets. In this context, any unplanned downtime due to equipment failure results in significant production losses, increased operational costs, and potential safety hazards for personnel \cite{huang2024prognostics}. To mitigate these risks, the paradigm is shifting from traditional preventive maintenance, which is often costly and fails to prevent all failures, towards Prognostics and Health Management (PHM). PHM aims to leverage the large volumes of data from the Industrial Internet of Things (IIoT) to monitor equipment health in real-time and predict failures before they occur.

A central task in PHM is the prediction of Remaining Useful Life (RUL), which estimates the time left before a component or system fails. However, a robust RUL prediction is fundamentally dependent on a preceding step: the construction of a reliable Health Indicator (HI). An HI is a quantitative metric that translates complex sensor data (e.g., temperature, vibration, current) into a simplified, monotonic representation of an asset's degradation state \cite{zhou2024systematic,zhou2025remaining}. Many researchers argue that decomposing the prognostics problem into two stages; first, constructing a robust HI, and second, predicting RUL from this HI; simplifies the model and enhances the understanding of the degradation behavior compared to "end-to-end" approaches that map raw data directly to RUL \cite{zhou2025remaining,li2024novel}. This two-step methodology allows for a clearer understanding of the non-linear degradation process and provides a more stable foundation for forecasting. This paper focuses on the first, foundational step: the construction of a HI for an industrial robot fleet.

\section{Problem Statement}

Despite the recent success of data-driven methods, particularly deep learning models like Autoencoders applied to HI construction \cite{mao2023new,li2024novel}, standard recon\-struc\-tion-based approaches face critical limitations in real-world industrial settings.

First, industrial robots exhibit highly "multi-modal" behavior. Here, multi-modality refers to the multi-modal distribution of a single univariate signal: the torque of one joint visits several disjoint operating regimes. In practice, the generated time series are discontinuous, jumping abruptly between distinct kinematic modes (e.g., waiting, fast movement, high-torque holding). If we observe these dynamics at a macro level, the data points form dense concentrations around specific valid states, separated by sparse transitional values. There is no strict temporal regularity or clear deterministic sequence between these modes, and the signals are subject to frequent temporal phase shifts. Furthermore, these operational modes lack explicit labels, making supervised segregation impossible. Standard regression-based models minimize the Mean Squared Error (MSE) over these sequences \cite{gonzalez2022health,liu2020complex}. Consequently, they tend to over-smooth the complex dynamics, converging towards an average representation of these disjointed modes. These methods are insensitive to mode deviations and therefore to early signs of degradation. They are only sensitive to extreme signal variations, which do not occur in all cases and occur late.

Second, deploying predictive maintenance across a fleet of thousands of robots introduces operational constraints regarding traceability and frugality. The "black-box" nature of deep learning models hinders their adoption by maintenance teams. Engineers require interpretable insights to trust the system's recommendations \cite{wagner2019implementing}: when an alarm is triggered, it must be traceable back to a specific kinematic state and physical reality. Moreover, deploying bespoke architectures per asset violates the frugality constraints of large-scale monitoring. An ideal solution should bypass specialized ML hardware (e.g., GPUs) and rely on operations simple enough to run directly within standard existing data architectures.

To address these limitations, we propose a methodological shift from sequential temporal reconstruction to geometric representation learning. The main contributions of this paper are:
\begin{itemize}
    \item We introduce a geometric HI framework utilizing Phase Space Reconstruction to handle multi-modal industrial dynamics.
    \item We provide a benchmark comparing discrete support estimation, deep learning models, and classical anomaly detection methods on the exact same phase space input, proving that boundary estimation outperforms reconstruction for degradation tracking in a "multi-modal" setup.
    \item We validate the proposed methodology on an industrial dataset and a synthetic multi-modal Langevin system to ensure reproducibility.
    \item We demonstrate that Discrete Support Estimation offers a pragmatic and easily deployable approach perfectly tailored to the realities of industrial constraints
\end{itemize}

\section{Related Work}

\subsection{Classical Statistical Approaches}
Before the advent of Deep Learning, statistical methods such as Change Point Detection (CPD) and Isolation Forests were widely explored for anomaly detection. 
However, their application to robotic fleets faces intrinsic limitations. 
Standard CPD algorithms (e.g., CUSUM) rely on stationarity assumptions that are violated by the highly non-stationary, multi-modal nature of robotic cycles, leading to excessive false alarms at every kinematic transition. 
Similarly, outlier detection methods like Isolation Forests typically isolate anomalies based on data sparsity. Consequently, they often conflate rare but entirely nominal operational modes (e.g., infrequent torque tasks) with mechanical anomalies, leading to a high false positive rate in multi-modal environments.

\subsection{Deep Learning for Condition Monitoring}
Recent data-driven approaches for HI construction heavily leverage deep learning. Autoencoders (AEs) and their variants are widely used for unsupervised feature learning and anomaly detection, where the reconstruction error often serves as the HI \cite{gonzalez2022health,liu2020complex}. For instance, advanced models like Tensor Autoencoders (TAE) \cite{mao2023new} or Spatiotemporal Convolutional Autoencoders (STECAE) \cite{li2024novel} have been proposed to better capture complex spatio-temporal dependencies. Variational Autoencoders (VAEs) have also been adapted, sometimes with constraints to enforce degradation trends \cite{qin2021unsupervised}. Recurrent architectures like LSTMs are also popular, either as part of an autoencoder (LSTM-AE) \cite{ye2021health} or for modeling temporal patterns directly.

Furthermore, standard Deep Learning models generally assume that training and test data share the same distribution \cite{yang2023novel,li2021self}. In a heterogeneous fleet, variations in robot types, payloads, and kinematic tasks induce a significant domain shift between assets. While advanced architectures (e.g., contrastive learning or generative adversarial networks) and Transfer Learning could theoretically bridge this gap or better capture multi-modal distributions \cite{mao2023new,li2024novel,qin2021unsupervised}, they typically require a massive, representative source domain to learn universal features. More importantly, developing, fine-tuning, and maintaining such complex models for each heterogeneous asset intrinsically violates the computational and frugality constraints of large-scale industrial deployments. In our industrial reality, aggregating data from thousands of robots into such a generalist base is currently prohibitive due to infrastructure maturity and storage costs. With access limited to a small subset of the fleet, relying on large-scale pre-training is not a viable strategy. This creates a strong need for asset-specific methods that self-calibrate without depending on a massive, centralized dataset.

\subsection{Prognostics and RUL Estimation}
The PHM field often decomposes prognostics into a two-step process: first, constructing a robust HI, and second, predicting the RUL from this HI \cite{zhou2024systematic,zhou2025remaining}. Many sophisticated methods focus on this two-stage pipeline. For example, some approaches use feature fusion via AEs or MLPs to build an HI, which is then fed into models like LSTMs or Bayesian frameworks for RUL prediction \cite{li2020shape,xiang2023cocktail}. Others use similarity-based methods, comparing a current HI trajectory to a database of run-to-failure curves \cite{xia2022multiscale,guo2022unsupervised}.

However, this paradigm still fundamentally relies on the availability of run-to-failure data to either train the RUL predictor or populate the similarity database \cite{chen2022deep,xia2022multiscale}. Many methods also require explicit labels for the start of degradation \cite{chen2022deep} or are designed for a single failure mode \cite{li2020shape}, which is unrealistic in complex industrial settings.

\subsection{Dynamical Systems in PHM}
The analysis of dynamical systems through phase space reconstruction (PSR) is a well-established field for characterizing complex systems from time series data \cite{takens2006detecting,kennel1992determining}. While its application in PHM is less mainstream than deep learning, several studies have successfully used it for fault diagnosis. Early works focused on identifying specific failure modes in rotating machinery by observing changes in the phase portrait's topology. More recently, metrics derived from phase space, such as local prediction errors \cite{luo2021improved,liu2023degradation}, have been used as health features. However, these approaches often focus on diagnosis rather than continuous monitoring, can be sensitive to outliers, or require operational condition labels for normalization.
In contrast to these methods that characterize trajectories, our methodology focuses on the continuous monitoring of the attractor's global geometric shape. By explicitly learning the spatial boundaries of the healthy manifold, we track the structural drift of the system over time. This shifts the objective from instantaneous fault classification to the construction of a HI suited for long-term prognostics.

\section{The Proposed Framework}

\subsection{Phase Space Representation}

The fundamental assumption of our framework is that the operational cycles of industrial robots are governed by deterministic physical laws, naturally converging toward a low-dimensional attractor. In practice, the full internal state is rarely accessible. We typically rely on univariate observations, such as motor torque, acting as a one-dimensional projection of the underlying dynamics. Following Takens's embedding theorem \cite{takens2006detecting}, the topological properties of the original state space can be reconstructed from these partial observations using a delayed embedding. We transform the discrete univariate time-series $\mathcal{X}=\{x_{1},x_{2},...,x_{N}\}$ into a multi-dimensional state vector $v_t$:

\begin{equation}
v_{t}=[x_{t},x_{t-\tau},...,x_{t-(m-1)\tau}]^{\top}\in\mathds{R}^{m}
\end{equation}

The reconstruction validity depends on the time delay $\tau$ and the embedding dimension $m$. While typically estimated via the Average Mutual Information (AMI) function and the False Nearest Neighbor (FNN) algorithm respectively \cite{wallot2018calculation,buzug1992optimal}, optimal parameters strictly depend on the operational context and the sampling frequency of the monitored system.

\subsection{Support Estimation}

Once the univariate time-series is successfully embedded into a multi-dimensional phase space, the subsequent step is to systematically characterize the nominal operational dynamics. In dynamical systems terminology, this translates to learning the support of the healthy attractor, representing the specific, bounded region of the phase space densely visited by the state vector $v_t$ during normal operations. Because industrial robotic tasks are multi-modal, this healthy attractor constitutes a complex geometric shape encompassing all valid kinematic states.

To rigorously encapsulate this geometric signature, various estimators can be deployed. Density estimation techniques, such as Gaussian Mixture Models (GMM), can model the continuous probability distribution across the manifold. Alternatively, boundary-learning algorithms like One-Class Support Vector Machines (OCSVM) or Isolation Forests construct a mathematical envelope around nominal observations. Another effective paradigm is direct, non-parametric estimation through discrete space partitioning, mapping continuous state vectors into a finite grid to form a binary occupancy map. 

It is crucial to distinguish this explicit geometric support estimation from the implicit distribution modeling performed by standard deep learning models. By operating directly on the spatial envelope rather than minimizing a sequential reconstruction error, explicit estimators are inherently protected against the temporal phase shifts and amplitude over-smoothing that plague recon\-struc\-tion-based models in multi-modal environments.

\subsection{Health Scoring}

The final stage of the proposed framework translates the geometric evaluation of the phase space into a continuous and actionable HI. For any new observation $v_t$ during the monitoring phase, the chosen support estimator outputs an instantaneous anomaly score $a_t$. Depending on the specific algorithm deployed within the framework, $a_t$ can manifest as a continuous metric or a discrete out-of-distribution flag; its concrete form is made explicit in Eq.~4 for our reference implementation and in Section~5 for each baseline.

However, raw point-wise anomaly scores are noisy and sensitive to transient variations: in multi-modal applications, a momentary deviation does not signify irreversible failure. To extract a reliable degradation trend, the instantaneous scores must be temporally aggregated.

We define the continuous Health Indicator, $HI(t)$, as the moving average of the anomaly scores over a sliding observation window of size $W$:

\begin{equation}
HI(t) = \frac{1}{W} \sum_{k=0}^{W-1} a_{t-k}
\end{equation}

The selection of the window size $W$ is a critical physical parameter. $W$ must satisfy the ergodic requirement of the monitored system: it must be sufficiently large to encapsulate multiple complete operational cycles. This ensures that the state vector $v_t$ has adequate time to densely and representatively sample the valid regions of the phase space.

\subsection{Reference Implementation: Discrete Support Estimation}

\begin{figure}[htbp]
\centering
\includegraphics[width=\textwidth]{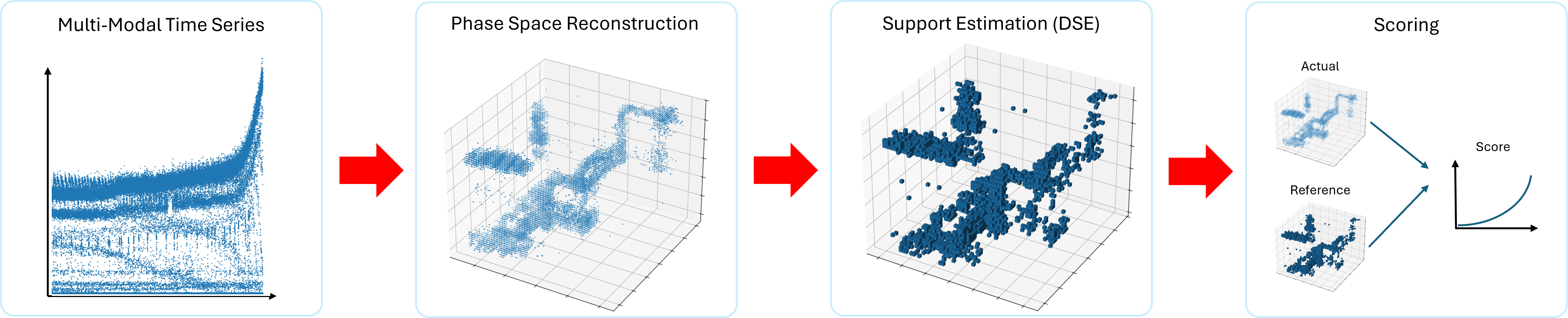}
\caption{Overview of the proposed geometric monitoring framework applied to Discrete Support Estimation.}
\label{fig:pipeline}
\end{figure}

While the framework accommodates various estimators, we propose (Fig.~\ref{fig:pipeline}) a non-parametric approach based on discrete space partitioning. This choice is motivated by the non-convex nature of multi-modal robotic attractors, which are naturally captured by a discrete grid. Furthermore, it yields a bounded binary score readily interpretable by domain experts, while guaranteeing strict computational frugality.

\paragraph{Discretization and Calibration}

During the calibration phase, we define the healthy signature using a reference dataset $\mathcal{D}_{ref}=\{v_{t}\}_{t=1}^{T_{cal}}$ observed during the robot's nominal period. We discretize the $m$-dimensional phase space into a regular grid with a resolution of $B$ bins per axis. A mapping function $\Phi:\mathds{R}^{m}\rightarrow\mathds{N}^{m}$ transforms each continuous state vector $v_t$ into discrete coordinates via min-max normalization and integer floor division. The healthy support $\mathcal{S}$ is then defined as the set of all unique cells visited during this period:

\begin{equation}
    \mathcal{S} = \{ \Phi(v) \mid v \in \mathcal{D}_{ref} \}
\end{equation}

In practice, $\mathcal{S}$ is implemented as a hash set, enabling $\mathcal{O}(1)$ insertion and lookup operations. The calibration process strictly requires a single pass over $\mathcal{D}_{ref}$, yielding an overall $\mathcal{O}(N)$ time complexity. Since the hash set stores only visited cells, the memory footprint is $\mathcal{O}(|\mathcal{S}|) \le \mathcal{O}(N)$, never the full $B^m$ grid. Raising $m$ thus costs coverage, not memory: for a fixed calibration budget the support grows sparse, making benign observations fall in unvisited cells (Sec.~\ref{sec:sensitivity_analysis}).

\paragraph{Anomaly Scoring and Health Indicator}

During the monitoring phase, structural degradation is identified when the system's dynamics drift into previously unvisited regions of the phase space. For each new observation $v_t$, a binary anomaly score $a_t$ is computed in constant time based on its membership in the healthy support:

\begin{equation}
    a_t = 
    \begin{cases} 
    0 & \text{if } \Phi(v_t) \in \mathcal{S} \\
    1 & \text{if } \Phi(v_t) \notin \mathcal{S}
    \end{cases}
\end{equation}

The final continuous Health Indicator is then obtained by passing this binary sequence $a_t$ through the ergodic sliding window described in Eq. (2).

We deliberately retain a binary membership test rather than a distance-based score (e.g., a truncated distance transform or an optimal-transport distance). For a non-convex, multi-modal support made of disjoint visited regions, the distance to the healthy set is geometrically ambiguous and, in our experiments, a poor degradation signal: what matters is whether the state has left the visited manifold, not how far one nominal mode lies from another.

\section{Experimental Setup}

\subsection{Industrial Dataset}

To validate the proposed framework under real-world constraints, we utilize a proprietary dataset extracted from an automotive manufacturing robot fleet comprising approximately 5000 industrial robots. For this study, a representative subset of 21 robots was selected. The historical data spans a period of up to three years of continuous operation.

The monitored condition variable is the torque disturbance, defined as the residual difference between the controller's reference command and the actual measured torque. Each robot consists of 6 distinct joints, yielding 6 independent and unsynchronized univariate time series per asset. As these joints are mechanically driven by separate gear trains and actuators, we treat each joint as an independent univariate channel: the framework is applied per joint, producing one health indicator per joint rather than a single joint-coupled embedding. This keeps the embedding low-dimensional (Sec.~\ref{sec:sensitivity_analysis}) and yields a localized indicator that points maintenance directly to the faulty axis.

Within this subset, 16 robots operated nominally without any documented mechanical issues throughout the three-year period. The remaining 5 robots experienced confirmed mechanical failures at known timestamps. For these specific assets, the time series are strictly truncated at the exact date of failure. Crucially, aligning with the unsupervised paradigm of our methodology, the data from these 5 run-to-failure trajectories is used exclusively for the final evaluation of the HI.

\subsection{Synthetic Dataset}

To replicate the multi-modal nature of industrial robotic cycles, we employ an over-damped, multi-stable Langevin equation. The state $x$ of the synthetic system evolves according to the following equation:

$$dx = (F_{phys}(x, w) + F_{control}(x, t))dt + \sigma dW_t$$

where $W_t$ is a standard Wiener process scaled by a noise magnitude $\sigma$. The proportional controller $F_{control}(x, t)$ drives the system toward a discrete cyclic target sequence, forcing continuous multi-modal transitions. The physical environment $F_{phys}(x, w)$ is modeled as a force derived from a non-linear potential field containing five distinct wells, representing the different valid operational modes:

$$F_{phys}(x, w) = \sum_{i=1}^{5} -k_i(w) (x - c_i(w)) \exp\left(-\frac{(x - c_i(w))^2}{0.5}\right)$$

To evaluate the HIs, we introduce a structural degradation parameterized by $w(t) \in [0, 1]$. For the first 40\% of the simulation, the system operates nominally. Subsequently, $w(t)$ increases linearly to $1.0$. This parameter alters the physical potential by inducing a spatial drift in specific wells (shifting the centers $c_i(w)$). Crucially, to simulate the integral action of a real robotic controller compensating for increased mechanical resistance, the target sequence of the proportional controller is symmetrically shifted.

\subsection{General Parameters and Phase Space Configuration}

To ensure a fair and rigorous comparative analysis, all evaluated baselines receive the exact same input dimension and are subjected to the same temporal aggregation. For the monitoring phase, all instantaneous anomaly scores are smoothed using an identical sliding window of size $W=5000$. This specific size guarantees the ergodicity of the evaluation by encompassing multiple operational cycles.

The Phase Space Reconstruction depends on the embedding dimension $m$ and the time delay $\tau$. For the synthetic Langevin system, the Average Mutual Information (AMI) function identifies an optimal delay at $\tau=58$, and the False Nearest Neighbors (FNN) algorithm confirms that the percentage of false neighbors drops to near zero at $m=3$. We retain $m=3$ as it fully unfolds the dynamics while preserving the ability to visually interpret the structural drift of the 3D attractor.

However, for the industrial dataset, AMI and FNN estimations were unstable across different robots. Consequently, we adopt a fixed configuration of $m=3$ and $\tau=1$ across the entire fleet. We chose this value for the delay $\tau$ because robots time series are already discontinuous. Indeed, the robotic joints are sampled at a low frequency (5 Hz) relative to their movements. The embedding dimension $m=3$ is chosen because the macroscopic movements of robots are inherently low-dimensional.

\begin{figure}[ht]
\centering
\includegraphics[width=\textwidth]{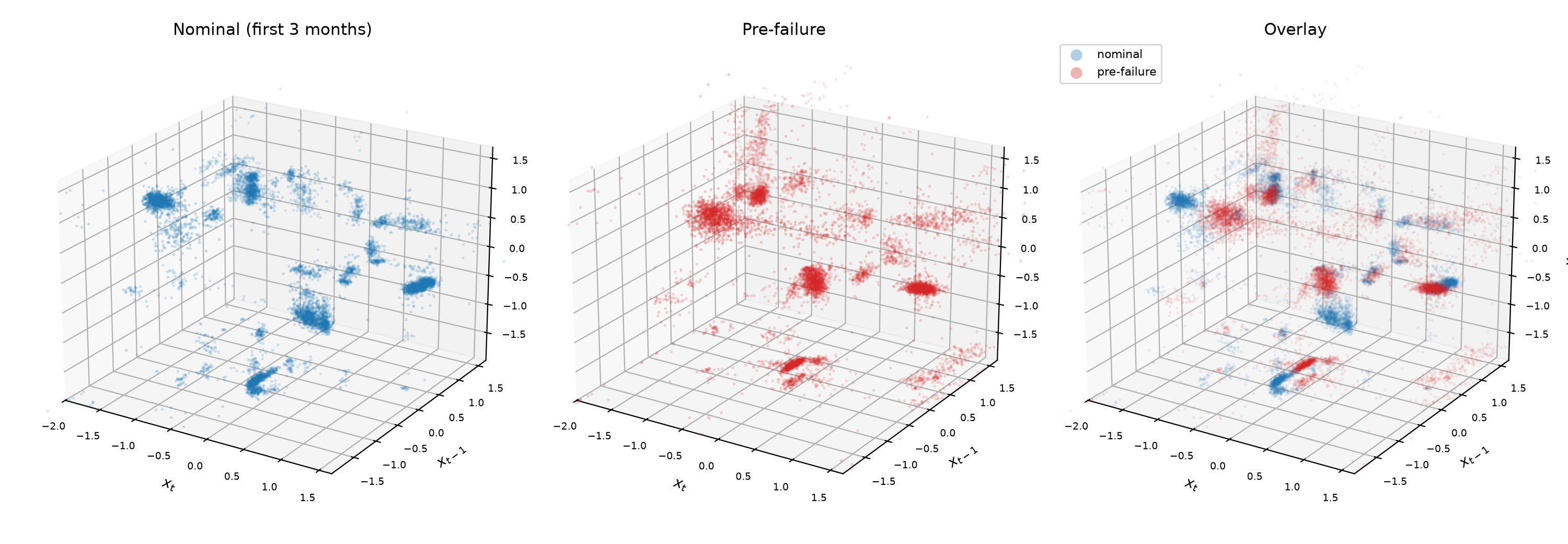}
\caption{Reconstructed phase-space attractor ($m=3$, $\tau=1$) for one joint of a failed robot. Left: nominal period (first three months); centre: pre-failure period; right: overlay. The fixed embedding yields a well-unfolded, multi-modal attractor, and degradation appears as the state populating cells outside the nominal support.}
\label{fig:attractor}
\end{figure}

To strictly validate this empirical configuration, a sensitivity analysis on the phase space parameters is included in Section~\ref{sec:sensitivity_analysis}. Figure~\ref{fig:attractor} shows the resulting industrial attractor, confirming that the fixed embedding is well-unfolded and multi-modal, with degradation appearing as the state drifting outside the nominal support.

\subsection{Baselines}

We benchmark our Discrete Support Estimation (DSE) against a comprehensive set of anomaly detection models. All baselines in the main experiment are fed the exact same 3D input vector ($v_t \in \mathds{R}^3$). For full reproducibility, the implementation of all baselines and the synthetic Langevin data generator are publicly available at \url{https://github.com/mart1bb/geometric-attractor-monitoring}.

\paragraph{Geometric and Statistical Models}
We evaluate several classical anomaly detection algorithms using the scikit-learn library:
\begin{itemize}
    \item \textbf{DSE:} Our reference implementation with a grid resolution fixed to $B=100$ bins per axis.
    \item \textbf{Gaussian Mixture Model:} Configured with $k=15$ components. The anomaly score is the negative log-likelihood of the sample.
    \item \textbf{One-Class SVM:} Deployed with a Radial Basis Function (RBF) kernel, scale gamma, and $\nu=0.01$.
    \item \textbf{Isolation Forest:} Configured with default parameters and an auto-adjusted contamination rate.
    \item \textbf{Centroid:} A baseline computing the Euclidean distance to the 3D center of mass of the healthy state vectors.
\end{itemize}

\paragraph{Deep Learning Models}
We implement standard deep reconstruction and forecasting architectures. All multi-layer perceptron (MLP) based encoders share a harmonized architecture of $[256, 128, 64, 32, d_{latent}]$, with symmetric decoders. All models are trained using the Mean Squared Error (MSE) loss, the Adam optimizer, and a learning rate of $10^{-4}$.
\begin{itemize}
    \item \textbf{Autoencoder (AE) and Variational Autoencoder (VAE):} The latent space dimension is strictly constrained to a structural bottleneck of $d_{latent}=2$. For the VAE, we extract two distinct anomaly scores: the sequential reconstruction error (VAE Rec) and the latent space divergence (VAE latent).
    \item \textbf{LSTM and LSTM-AE:} Recurrent models configured with two stacked hidden layers of 64 units.
    \item \textbf{MLP Predictor:} A feed-forward network forecasting the next coordinate $x_{t+1}$.
\end{itemize}

\subsection{Evaluation Metrics}
\label{subsec:metrics}

To objectively evaluate the performance of the generated HI, we formalize a comprehensive evaluation framework based on standard metrics widely adopted in the PHM literature. Let $\mathbf{h} = (h_1, \dots, h_K)$ denote the sequence of HI values for a given operational trajectory of length $K$, and $\mathbf{t} = (1, \dots, K)$ be the corresponding time vector.

\paragraph{Monotonicity}
Monotonicity measures the global irreversibility of the wear process regardless of its non-linear functional form (e.g., exponential degradation). To robustly handle local noise and signal length variations without requiring explicit smoothing, we define Monotonicity as the absolute Spearman rank correlation coefficient between the HI and time:
\begin{equation}
    \text{Mon} = \left| \frac{\sum_{t=1}^{K} (\text{rank}(h_t) - \overline{\text{rank}(\mathbf{h})})(\text{rank}(t) - \overline{\text{rank}(\mathbf{t})})}{\sqrt{\sum_{t=1}^{K} (\text{rank}(h_t) - \overline{\text{rank}(\mathbf{h})})^2} \sqrt{\sum_{t=1}^{K} (\text{rank}(t) - \overline{\text{rank}(\mathbf{t})})^2}} \right|
\end{equation}

\paragraph{Temporal Correlation}
While Monotonicity evaluates rank-based irreversibility, Temporal Correlation strictly quantifies the linearity of the degradation rate. This evaluates the HI's suitability for direct linear RUL extrapolation. We compute this using the absolute Pearson correlation coefficient:
\begin{equation}
    \text{Corr} = \left| \frac{\sum_{t=1}^{K} (h_t - \bar{h})(t - \bar{t})}{\sqrt{\sum_{t=1}^{K} (h_t - \bar{h})^2} \sqrt{\sum_{t=1}^{K} (t - \bar{t})^2}} \right|
\end{equation}

\paragraph{Amplitude-Normalized Robustness}
Robustness measures the stability of the indicator relative to its own dynamic range. To isolate the macroscopic degradation trend from local residual noise, we extract a perfectly smoothed baseline $h_t^{trend}$ using a Savitzky-Golay filter (polynomial order 2, window length $\approx K/5$). The robustness is then computed by penalizing the residuals normalized by the total degradation amplitude:
\begin{equation}
    \text{Rob} = \exp \left( - \frac{1}{K} \sum_{t=1}^{K} \left| \frac{h_t - h_t^{trend}}{\max(\mathbf{h}) - \min(\mathbf{h})} \right| \right)
\end{equation}

\paragraph{Prognosticability}
Prognosticability evaluates the consistency of the final HI value at the moment of failure across multiple assets. A reliable HI should trigger an alarm at a similar threshold regardless of the specific robot. Let $\mathbf{h}_{end}$ be the vector of the final HI values for the broken population, and $\Delta \mathbf{h}$ be the vector of total degradation amplitudes for these assets.
\begin{equation}
    \text{Prog} = \exp \left( - \frac{\text{std}(\mathbf{h}_{end})}{\text{mean}(\Delta \mathbf{h})} \right)
\end{equation}

\paragraph{Discriminability (Signal-to-Noise Proxy)}
In industrial deployments, minimizing false alarms is a strict constraint. Leveraging our control group of nominal robots, we introduce a Discriminability metric to quantify the separation between true degradation signatures and operational baseline noise.
\begin{equation}
    \text{Disc} = \frac{1}{1 + \frac{\bar{\sigma}_{healthy}}{\bar{\Delta}_{broken}}}
\end{equation}
where $\bar{\sigma}_{healthy}$ is the average standard deviation of the HIs observed on the healthy fleet, and $\bar{\Delta}_{broken}$ is the average degradation amplitude of the failed robots. A score close to 1 indicates that the degradation is distinctively separated from operational variability.

\paragraph{Computational Efficiency}
Finally, to validate the frugality constraint of monitoring, we measure the end-to-end execution time.

\section{Results and Comparative Analysis}

\subsection{Overall Monitoring Performance}

\begin{figure}[htbp]
\centering
\begin{minipage}{0.48\textwidth}
\centering
\includegraphics[width=\linewidth]{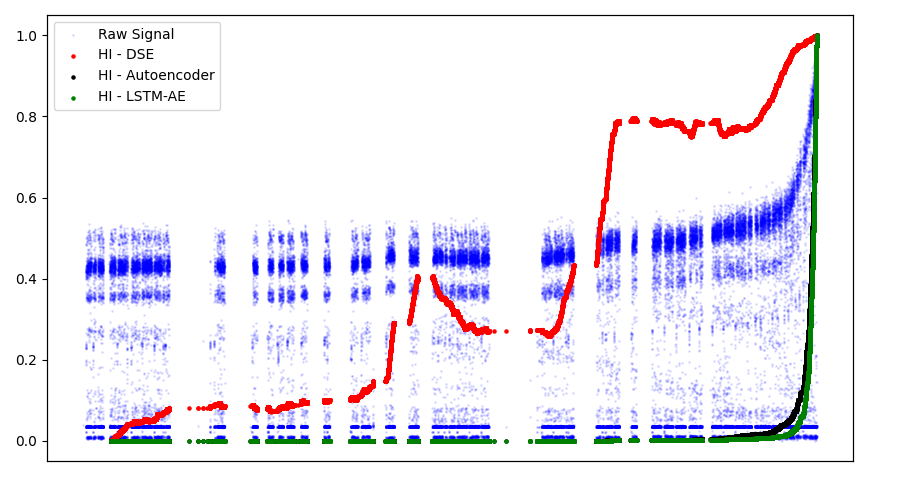}
\caption{Comparison of normalized indicators for Robot 1 that had breakdown}
\label{fig:plot1}
\end{minipage}\hfill
\begin{minipage}{0.48\textwidth}
\centering
\includegraphics[width=\linewidth]{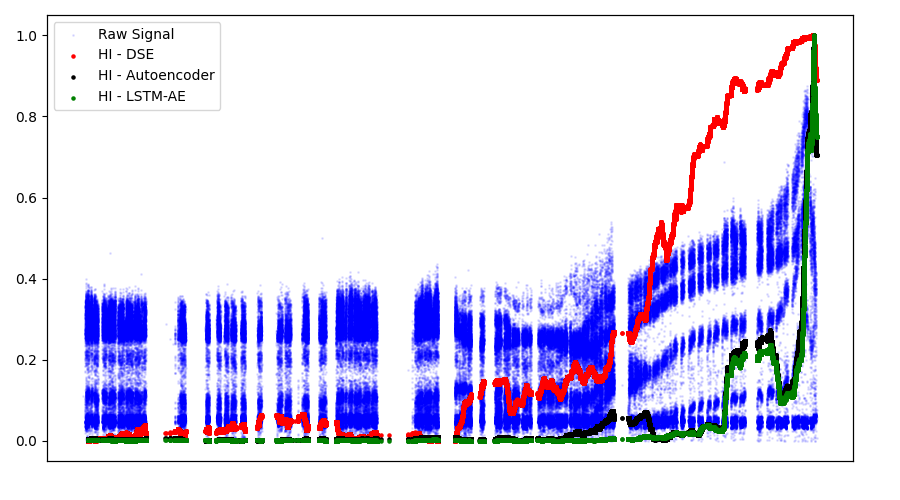}
\caption{Comparison of normalized indicators for Robot 2 that had breakdown}
\label{fig:plot2}
\end{minipage}
\end{figure}

Table \ref{tab:global_results} presents the quantitative evaluation of the HIs generated by all baseline models on the industrial dataset evaluated on the exact same low-dimensional phase space input ($v_t \in \mathds{R}^3$). Figures~\ref{fig:plot1} and~\ref{fig:plot2} illustrate the resulting indicators on two failed robots: DSE exhibits a clear rising trend toward failure, while recon\-struc\-tion-based baselines remain comparatively flat. As these traces are not strictly monotonic, we rely on the rank-based Monotonicity (Eq.~5), which captures the global degradation trend while tolerating local fluctuations.

\begin{table}[ht]
\centering
\caption{Overall monitoring performance on the industrial fleet}
\label{tab:global_results}
\begin{tabular}{@{}lccccccc@{}}
\toprule
\textbf{Method} & \textbf{Mon.} & \textbf{Time Corr.} & \textbf{Rob.} & \textbf{Prog.} & \textbf{Disc.} & \textbf{Global} & \textbf{Time(s)} \\ \midrule
DSE & $\mathbf{0.61} {\scriptstyle \pm 0.20}$ & $\mathbf{0.66} {\scriptstyle \pm 0.13}$ & $0.94 {\scriptstyle \pm 0.02}$ & \textbf{0.54} & 0.77 & \textbf{0.70} & $0.28 {\scriptstyle \pm 0.34}$ \\
GMM & $0.59 {\scriptstyle \pm 0.15}$ & $0.62 {\scriptstyle \pm 0.09}$ & $\mathbf{0.95} {\scriptstyle \pm 0.02}$ & 0.35 & 0.81 & 0.66 & $3.55 {\scriptstyle \pm 5.83}$ \\
IF & $0.50 {\scriptstyle \pm 0.15}$ & $0.54 {\scriptstyle \pm 0.11}$ & $0.93 {\scriptstyle \pm 0.05}$ & 0.45 & 0.73 & 0.63 & $0.60 {\scriptstyle \pm 0.58}$ \\
OCSVM & $0.51 {\scriptstyle \pm 0.08}$ & $0.50 {\scriptstyle \pm 0.11}$ & $0.93 {\scriptstyle \pm 0.03}$ & 0.23 & \textbf{0.91} & 0.62 & $79.09 {\scriptstyle \pm 244.94}$ \\
Centroid & $0.52 {\scriptstyle \pm 0.08}$ & $0.53 {\scriptstyle \pm 0.06}$ & $0.93 {\scriptstyle \pm 0.03}$ & 0.35 & 0.67 & 0.60 & $\mathbf{0.01} {\scriptstyle \pm 0.01}$ \\
VAE (Rec) & $0.52 {\scriptstyle \pm 0.12}$ & $0.52 {\scriptstyle \pm 0.12}$ & $0.93 {\scriptstyle \pm 0.02}$ & 0.32 & 0.72 & 0.60 & $3.04 {\scriptstyle \pm 3.36}$ \\
LSTM-AE & $0.47 {\scriptstyle \pm 0.15}$ & $0.45 {\scriptstyle \pm 0.14}$ & $0.92 {\scriptstyle \pm 0.03}$ & 0.28 & 0.73 & 0.57 & $3.19 {\scriptstyle \pm 3.59}$ \\
LSTM (Pred) & $0.44 {\scriptstyle \pm 0.06}$ & $0.50 {\scriptstyle \pm 0.04}$ & $0.92 {\scriptstyle \pm 0.03}$ & 0.24 & 0.76 & 0.57 & $6.16 {\scriptstyle \pm 7.53}$ \\
MLP & $0.48 {\scriptstyle \pm 0.11}$ & $0.48 {\scriptstyle \pm 0.11}$ & $0.93 {\scriptstyle \pm 0.02}$ & 0.15 & 0.64 & 0.54 & $2.40 {\scriptstyle \pm 2.83}$ \\
Autoencoder & $0.42 {\scriptstyle \pm 0.12}$ & $0.45 {\scriptstyle \pm 0.10}$ & $0.93 {\scriptstyle \pm 0.03}$ & 0.03 & 0.47 & 0.46 & $2.77 {\scriptstyle \pm 3.11}$ \\
VAE (latent) & $0.49 {\scriptstyle \pm 0.14}$ & $0.48 {\scriptstyle \pm 0.11}$ & $0.93 {\scriptstyle \pm 0.02}$ & 0.01 & 0.11 & 0.40 & $3.04 {\scriptstyle \pm 3.36}$ \\ \bottomrule
\end{tabular}
\end{table}

While continuous estimators like GMM provide competitive local trend tracking, their high variance across assets reveals a sensitivity to fleet heterogeneity. DSE achieves the highest global score by converting the continuous state into discrete grid boundaries, which immunizes it against the cyclic fluctuations that penalize MSE-based models. This directly translates into higher Prognosticability: the accumulation of strict binary boundary violations keeps the final degradation amplitude consistent across the fleet, whereas unconstrained MSE values vary wildly per asset, resulting in poor prognostic consistency.

While some baselines exhibit higher Discriminability (e.g., OCSVM or AE), this is a common artifact of overly conservative models that output a flat signal across both healthy and broken phases, artificially inflating the signal-to-noise ratio proxy. Their failure to capture the actual degradation renders them unsuitable for predictive maintenance. DSE provides the optimal operational trade-off, successfully tracking the structural drift while maintaining a high rejection rate for false alarms.

We stress that Prognosticability and Discriminability are aggregated over the $N=5$ run-to-failure assets available in the fleet, so these fleet-level scores carry non-negligible uncertainty. To quantify it, we report bootstrap confidence intervals over the failed population in the supplementary material; DSE retains the highest Prognosticability in 96\% of resamples, though the wide intervals confirm that the absolute values should be read as indicative rather than precise.

Beyond diagnostic accuracy, Table \ref{tab:global_results} shows that DSE processes a complete trajectory with marked computational frugality, allowing the framework to scale across the fleet without specialized hardware acceleration, avoiding the computational overhead of standard deep learning architectures or classical kernel-based methods.

\begin{table}[ht]
\centering
\caption{Monitoring performance on the synthetic multi-stable Langevin system}
\label{tab:langevin_results}
\begin{tabular}{@{}lcccc@{}}
\toprule
\textbf{Method} & \textbf{Mon.} & \textbf{Time Corr.} & \textbf{Rob.} & \textbf{Global Score} \\ \midrule
Isolation Forest & 0.996 & \textbf{0.995} & 0.982 & \textbf{0.991} \\
DSE              & \textbf{0.998} & 0.983 & \textbf{0.989} & 0.990 \\
GMM              & 0.998 & 0.984 & 0.986 & 0.989 \\
Centroid         & 0.991 & 0.991 & 0.973 & 0.985 \\
VAE (latent)     & 0.990 & 0.989 & 0.969 & 0.983 \\
Autoencoder      & 0.986 & 0.986 & 0.963 & 0.978 \\
OCSVM            & 0.965 & 0.966 & 0.939 & 0.957 \\
MLP              & 0.789 & 0.794 & 0.932 & 0.838 \\
VAE (Rec)        & 0.724 & 0.740 & 0.896 & 0.787 \\
LSTM-AE          & 0.716 & 0.746 & 0.897 & 0.786 \\
LSTM (Pred)      & 0.593 & 0.627 & 0.858 & 0.693 \\ \bottomrule
\end{tabular}
\end{table}

Table \ref{tab:langevin_results} presents the monitoring performance on the synthetic Langevin system. These results corroborate the findings from the industrial dataset: geometric and spatial bounding estimators provide a viable alternative to, or even perform better than deep learning baselines in "multi-modal" environment while remaining more computationally efficient. A notable difference with the industrial evaluation is that continuous estimators like Isolation Forest marginally outperform DSE on this synthetic benchmark. This inversion is structurally logical: the injected degradation is a perfectly continuous spatial drift, which continuous estimators accommodate better than a rigid discrete grid. In real-world industrial settings, however, transient noise and non-linear shocks favor the strict spatial boundaries of DSE.

\subsection{Long Context Window}

\begin{table}[ht]
\centering
\caption{Deep learning models with extended context window}
\label{tab:ablation_results}
\begin{tabular}{@{}lccccccc@{}}
\toprule
\textbf{Method} & \textbf{Mon.} & \textbf{Time Corr.} & \textbf{Rob.} & \textbf{Prog.} & \textbf{Disc.} & \textbf{Global} & \textbf{Time(s)} \\ \midrule
CNN-AE & $0.45 {\scriptstyle \pm 0.20}$ & $0.53 {\scriptstyle \pm 0.12}$ & $0.93 {\scriptstyle \pm 0.02}$ & \textbf{0.30} & 0.72 & \textbf{0.59} & $170.04 {\scriptstyle \pm 250.97}$ \\
VAE (Rec) & $0.45 {\scriptstyle \pm 0.19}$ & $\mathbf{0.53} {\scriptstyle \pm 0.11}$ & $\mathbf{0.93} {\scriptstyle \pm 0.01}$ & 0.27 & 0.73 & 0.58 & $8.62 {\scriptstyle \pm 10.36}$ \\
LSTM-AE & $0.42 {\scriptstyle \pm 0.22}$ & $0.49 {\scriptstyle \pm 0.13}$ & $0.91 {\scriptstyle \pm 0.03}$ & 0.23 & 0.86 & 0.58 & $33.17 {\scriptstyle \pm 43.18}$ \\
LSTM (Pred) & $0.42 {\scriptstyle \pm 0.22}$ & $0.52 {\scriptstyle \pm 0.14}$ & $0.91 {\scriptstyle \pm 0.03}$ & 0.24 & 0.80 & 0.58 & $29.07 {\scriptstyle \pm 37.76}$ \\
Autoencoder & $\mathbf{0.46} {\scriptstyle \pm 0.19}$ & $0.51 {\scriptstyle \pm 0.13}$ & $0.93 {\scriptstyle \pm 0.02}$ & 0.28 & 0.72 & 0.58 & $\mathbf{8.10} {\scriptstyle \pm 9.96}$ \\
VAE (latent) & $0.44 {\scriptstyle \pm 0.23}$ & $0.46 {\scriptstyle \pm 0.08}$ & $0.93 {\scriptstyle \pm 0.03}$ & 0.12 & \textbf{0.86} & 0.56 & $8.62 {\scriptstyle \pm 10.36}$ \\
MLP & $0.35 {\scriptstyle \pm 0.13}$ & $0.50 {\scriptstyle \pm 0.14}$ & $0.93 {\scriptstyle \pm 0.02}$ & 0.25 & 0.74 & 0.55 & $11.53 {\scriptstyle \pm 12.01}$ \\
Transformer & $0.37 {\scriptstyle \pm 0.15}$ & $0.35 {\scriptstyle \pm 0.10}$ & $0.91 {\scriptstyle \pm 0.01}$ & 0.07 & 0.55 & 0.45 & $274.97 {\scriptstyle \pm 404.19}$ \\ \bottomrule
\end{tabular}
\end{table}

To verify if the underperformance of deep learning models stems from an insufficient temporal receptive field, we conducted a study by expanding their context window. The models were provided with an input sequence of $W=5000$ time steps, perfectly matching the ergodic window, with the exception of the recurrent architectures (LSTM and LSTM-AE) which were restricted to $W=200$ due to memory constraints. We relaxed the latent bottleneck ($d_{latent}=20$) and introduced two other sequence-to-sequence architectures: a 1D Convolutional Autoencoder (CNN-AE) and a Transformer.

The results, presented in Table \ref{tab:ablation_results}, confirm that increasing the context window yields only marginal improvements: the deep architectures consistently fail to match the frugal DSE estimator. This persistent underperformance, despite a receptive field matching the ergodic window, confirms that the bottleneck is the architectural over-smoothing bias discussed in Sec.~2, not an insufficient temporal context.

\subsection{Sensitivity Analysis of Phase Space Parameters}
\label{sec:sensitivity_analysis}

\begin{figure}[htbp]
\centering
\begin{minipage}{0.43\textwidth}
\centering
\includegraphics[width=\linewidth]{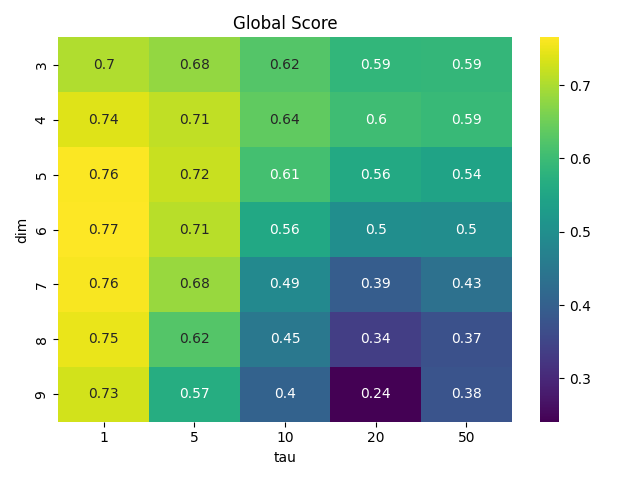}
\caption{Global Score as a function of dimension $m$ and delay $\tau$}
\label{fig:parameter_matrix}
\end{minipage}\hfill
\begin{minipage}{0.55\textwidth}
\centering
\includegraphics[width=\linewidth]{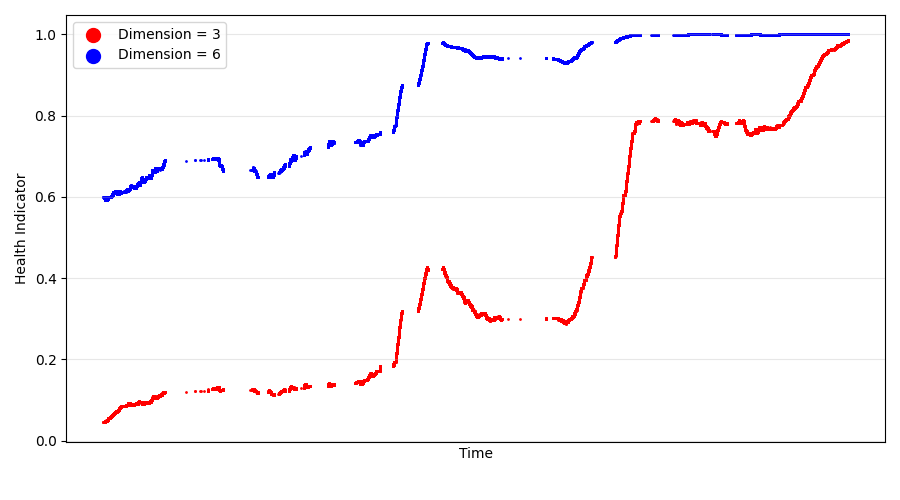}
\caption{Comparison of indicator calculated with dimension equal to 3 and 6}
\label{fig:indicator_comparison}
\end{minipage}
\end{figure}

To empirically validate the phase space configuration, we conducted a sensitivity analysis on the DSE framework by varying the embedding dimension $m$ and the time delay $\tau$. 
The experimental results (Fig.~\ref{fig:parameter_matrix}) demonstrate that the impact of the embedding dimension $m$ is dependent on the chosen time delay $\tau$.
For delays $\tau > 5$, increasing $m$ degrades the global performance score.
Conversely, for short delays ($\tau \le 5$), increasing the dimension artificially improves the overall score.
This phenomenon is the curse of dimensionality acting on the discrete support: as $m$ grows, the $B^m$ cells vastly outnumber the fixed calibration budget, so the healthy support becomes sparse and even benign observations fall in unvisited cells. This inflates prognosticability and causes the HI to saturate prematurely into a binary out-of-bounds trigger.
Although higher dimensions yield higher quantitative scores at $\tau \le 5$, we retain the configuration $m=3$.
This specific dimensionality prevents early saturation, yielding a continuous indicator that effectively tracks the physical degradation trend rather than acting as a simple binary out-of-bounds trigger (Fig.~\ref{fig:indicator_comparison}).
Furthermore, maintaining $m=3$ keeps the healthy support densely sampled and preserves the ability to visually interpret the structural drift of the 3D attractor.

\section{Discussion and Conclusion}

This paper presented a frugal and scalable geometric framework for the predictive maintenance of large-scale robotic fleets. By shifting the anomaly detection paradigm from temporal reconstruction to Phase Space Reconstruction, the Discrete Support Estimation (DSE) method isolates diverse kinematic regimes into discrete geometric regions without averaging them. Empirically, DSE outperforms deep architectures that suffer from an over-smoothing bottleneck on multi-modal cycles, while its geometric alerts remain directly traceable to a physical reality by maintenance engineers.

While more complex architectures might mitigate this bottleneck, maintaining such bespoke models per asset violates the frugality constraint of fleet-scale deployment; the geometric framework thus offers a pragmatic compromise. A primary limitation of the current DSE approach is its rigid reliance on a static nominal trajectory; if a robot is reprogrammed to perform a new operational task, the historical support becomes obsolete. Future work will focus on integrating continual learning mechanisms to dynamically update the discrete support, enabling the framework to autonomously adapt to industrial trajectory reprogramming without requiring full recalibration.

Additionally, ongoing research evaluates this geometric representation on other manufacturing assets subject to similar multi-modal characteristics, such as stamping presses, to further validate the generalizability of the proposed framework.

\subsubsection*{Disclosure of LLM usage.} During the preparation of this manuscript, the authors used a large language model to assist with language editing, text condensation, and LaTeX formatting. All scientific content, experimental design, and conclusions are the authors' own, and the authors take full responsibility for the final manuscript.

\bibliographystyle{splncs04}
\bibliography{references}

\end{document}